\documentclass[sigconf, 10pt]{acmart}
\usepackage{graphicx}
\usepackage{subcaption}
\usepackage{enumitem}
\usepackage{adjustbox,xspace}
\usepackage{siunitx}
\usepackage{xcolor}
\usepackage{multirow}
\usepackage{ragged2e}
\usepackage{tabularx}

\definecolor{mygreen}{RGB}{0, 204, 102}

\AtBeginDocument{%
  }
\renewcommand\footnotetextcopyrightpermission[1]{}  
\setcopyright{acmlicensed}
\copyrightyear{2018}
\acmYear{2018}
\acmDOI{XXXXXXX.XXXXXXX}
\acmConference[MobiCom'26]{}{Nov, 2026}{Austin, Texas, USA}

\acmISBN{978-1-4503-XXXX-X/2018/06}

\newcommand{\ie}{{\itshape i.e.}\xspace}
\newcommand{\eg}{\emph{e.g.}\xspace}
\newcommand{\etc}{\emph{etc.}\xspace}

\newcommand{\aka}{\emph{a.k.a.}\xspace}

\newcommand{\sysname}{\textsc{AURA}\xspace}

\begin{document}

\begin{teaserfigure}
\centering
{\includegraphics[width=1\linewidth]{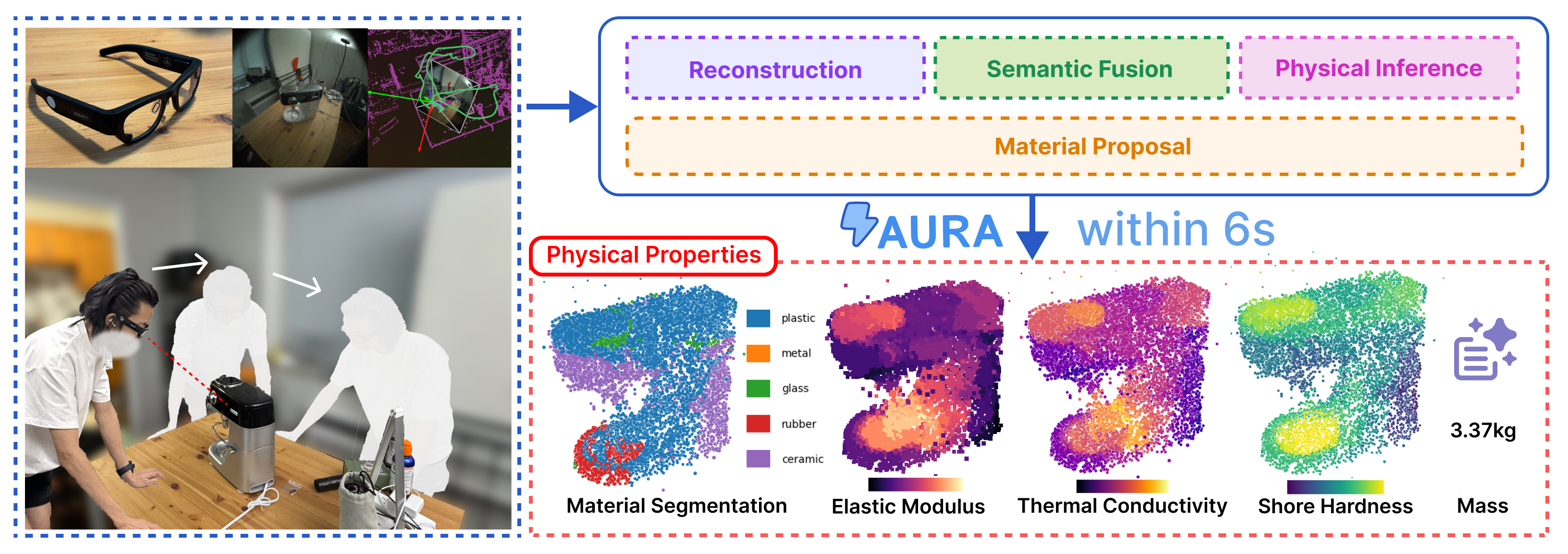}}
\vspace{-8mm}
\caption{A mobile user wearing Meta Aria  Glasses uses \sysname to estimate an object's weight and infer its physical properties like hardness, material distribution, and elasticity, all within 6 seconds. }
\label{fig:teaser}    
\end{teaserfigure}

\title[Accelerating Physical Property Reasoning for Augmented Visual Cognition]{
{Accelerating Physical Property Reasoning for Augmented Visual Cognition}}

\author{Hongbo Lan}
\affiliation{
  \institution{University of Pittsburgh}
  \city{Pittsburgh}
  \state{PA}
  \country{USA}
}
\email{hol101@pitt.edu}

\author{Zhenlin An}
\affiliation{
  \institution{University of Georgia}
  \city{Athens}
  \state{GA}
  \country{USA}
}
\email{zhenlin.an@uga.edu}

\author{Haoyu Li}
\affiliation{
  \institution{University of Pittsburgh}
  \city{Pittsburgh}
  \state{PA}
  \country{USA}
}
\email{hal308@pitt.edu}

\author{Vaibhav Singh}
\affiliation{
  \institution{University of Pittsburgh}
  \city{Pittsburgh}
  \state{PA}
  \country{USA}
}
\email{vaibhav.s@pitt.edu}

\author{Longfei Shangguan}
\affiliation{
  \institution{University of Pittsburgh}
  \city{Pittsburgh}
  \state{PA}
  \country{USA}
}
\email{longfei@pitt.edu}

\renewcommand{\shortauthors}{anonymous submission \#244}

\begin{abstract}
This paper introduces \sysname, a system that accelerates vision-guided physical property reasoning to enable augmented visual cognition. 
\sysname minimizes the run-time latency of this reasoning pipeline through a combination of both algorithmic and systematic optimizations, including rapid geometric 3D reconstruction, efficient semantic feature fusion, and parallel view encoding.
Through these simple yet effective optimizations, \sysname reduces the end-to-end latency of this reasoning pipeline from 10--20 minutes to less than 6 seconds. A head-to-head comparison on the ABO dataset shows that \sysname achieves this 62.9$\times$--287.2$\times$ speedup while not only reaching on-par (and sometimes slightly better) object-level physical property estimation accuracy(e.g. mass), but also demonstrating superior performance in material segmentation and voxel-level inference than two SOTA baselines.
We further combine gaze-tracking with \sysname to localize the object of interest in cluttered, real-world environments, streamlining the physical property reasoning on smart glasses. 
The case study with Meta Aria Glasses conducted at an IKEA furniture store demonstrates that \sysname achives consistently high performance compared to controlled captures, providing robust property estimations even with fewer views in real-world scenarios. 
Artifacts and demo will be released before publication at \textcolor{red}{\url{https://hl-demo.github.io/}}.

\end{abstract}

\maketitle

\section{Introduction}

As humans, we interact with objects in our daily life, from lifting bags to choosing groceries.
These interactions are largely guided by our visual cognition of an object's physical properties, such as its weight, texture, and material. For instance, we visually estimate a suitcase's weight before lifting it to prevent injury. When stacking dishes, we assess their fragility and stability to avoid breaking them. Similarly, we quickly gauge a chair's sturdiness before sitting down. This ability to intuitively assess an object's properties allows us to plan and execute our actions efficiently and safely.

However, visual cognition is often subject to individual biases, which are influenced by a person's experience, age, and cognitive ability ~\cite{wang2022individual,wexler2022structure}. 
For example, two people might look at the same chair and have vastly different ideas of its weight: one might perceive it as lightweight due to its design, while another might assume it's heavy based on its material.
These unconscious biases can lead to significant issues, \eg, an incorrect assessment of a heavy load's stability could cause a structural failure, or misjudging a tool's fragility could result in damage. Therefore, a mobile system that can accurately reason about an object's physical properties has the potential to complement and augment human visual cognition.
Below we list a few examples, among many others.

\noindent $\bullet$ \textbf{Logistics and Supply Chain}: During truck loading, such a system can automatically estimate the weight and balance of each item based on visual cues, enabling more accurate and efficient load planning. This helps ensure proper weight distribution across the vehicle without relying on time-consuming manual weighing.
    
\noindent $\bullet$ \textbf{Elderly Assistance}. People with cognitive challenges (\eg, dementia, traumatic brain injury, or intellectual disabilities) may struggle to judge whether an object is safe to lift. The system could function like a cognitive prosthetic, filling in for judgment or memory deficits.

\noindent $\bullet$ \textbf{Augmented Reality (AR) and Gaming}. Such a system can enhance object interactions in AR games by factoring in estimated weight for dynamics. Likewise, it can also be used in training scenarios (\eg, lifting safety) where realistic object properties matter.

Recently, two emerging trends, one driven by industry and the other by academia, are accelerating the realization of such an {\it augmented human cognition} system.
On the industrial front, the proliferation of smart glasses equipped with RGB cameras (\eg, Ray-Ban Meta Glasses~\cite{meta_glasses}) has made it increasingly feasible to capture and sense everyday objects seamlessly. In academia, the rapid advancement of vision-based 3D reconstruction techniques (\eg, NeRF~\cite{mildenhall2021nerf}, 3D Gaussian Splatting~\cite{kerbl20233d}), combined with the growing capabilities of vision-language models (VLMs)~\cite{radford2021learning,li2022blip,alayrac2022flamingo,zhu2023minigpt}, now enable the reconstruction of everyday objects and reasoning their physical properties such as weight, texture, and material.

While prior works~\cite{liu2025generative,zhai_physical_2024,shuai_pugs_2025} have explored VLM-driven physical property reasoning, these solutions often overlook a critical system-level consideration necessary for real-world augmented visual cognition applications -- {\it runtime latency}.
For instance, pioneer works such as NeRF2Physics~\cite{zhai_physical_2024} and PUGS~\cite{shuai_pugs_2025} typically require 10 to 20 minutes to infer the geometry, materials, and the weight of a single object (\S\ref{ss:3Dreasoning}). 
This means a user wearing smart glasses would have to take multiple photos of an object and then wait 10 to 20 minutes before the system can return the object's physical properties. 
This substantial delay makes them impractical for augmented visual cognition, where real-time performance is crucial for a seamless user experience.

\noindent This paper revisits existing VLM-driven physical property reasoning pipelines from a new holistic systems perspective. Instead of designing new vision algorithms, we carefully analyze each design component of this multi-stage pipeline to identify its key computation bottlenecks (\S\ref{ss:3Dreasoning}). 
Based on the insights from this analysis, we introduce \sysname (\textbf{A}ccelerated \textbf{U}nderstanding \& \textbf{R}easoning \textbf{A}nalyzer), the first system designed to minimize these processing delays through a combination of both algorithmic and systematic optimizations, marking a significant step toward real-time augmented visual cognition.

At the core of \sysname is a hierarchical optimization framework built on a key insight: reasoning about an object's physical properties does not require a computationally intensive, photo-realistic 3D reconstruction like those from NeRF and 3DGS \cite{zhai_physical_2024, shuai_pugs_2025}. Our framework achieves a significant reduction in the delay of multi-view physical property reasoning through a set of innovations summarized below.

\noindent $\bullet$ \textbf{One-Shot Reconstruction}: We use VGGT~\cite{wang2025vggt}, an end-to-end, one-shot model, to generate an object's 3D point cloud in a single forward pass. This bypasses the minutes of latency introduced by training-based methods adopted by SOTA solutions \cite{zhai_physical_2024, shuai_pugs_2025}.

\noindent $\bullet$ \textbf{Efficient Semantic Mapping and Filtering}: To overcome the unstructured nature of VGGT's point cloud, we leverage the rich DINO features that VGGT already generates internally to bridge the 3D point cloud and the multi-view object photos. By "tapping into" these DINO features, the system performs a semantically-aware adaptive downsampling and view-importance filtering. This drastically reduces the number of time-consuming semantic feature calls, while effectively clustering points from the same object component. Finally, we propose a multi-scale feature fusion algorithm to capture both local texture details and broader geometric context, making the final contextual representation more robust to object scale variation and viewpoint changes.

\noindent $\bullet$ \textbf{Parallelizing Semantic Feature Fusion}: We transform the semantic feature fusion pipeline from a prohibitively slow, serial process into a highly efficient parallel algorithm. Our proposed algorithm uses vectorized operations to simultaneously project all source points onto every view and batch-compute visibility and normal angle scores. This design allows us to select optimal patches and aggregate them into a single global batch for unified processing by the CLIP encoder, thereby dramatically reducing latency.

Through these simple yet effective optimizations, \sysname successfully reduces the end-to-end latency of physical property reasoning from 10-20 minutes to less than 6 seconds, marking a big step towards augmented visual cognition.
The head-to-head comparison with NeRF2Physics \cite{zhai_physical_2024} and PUGS \cite{shuai_pugs_2025}, two SOTA physical property reasoning works on gold-standard ABO dataset \cite{collins_abo_2022} demonstrate that \sysname achieves 62.9$\times$ to 287.2$\times$ delay reduction while maintaining accuracy competitive with NeRF2Physics and significantly better than PUGS on weight estimation accuracy.
To validate our system's efficiency in cluttered, real-world environments, we further propose to leverage gaze tracking to infer a user's intention and localize the object of interest without explicit human intervention. We integrate this component into \sysname and conduct a case study using Meta Aria smart glasses at an IKEA furniture store and lab environment. The results show that \sysname achieves only a slight decrease in mass estimation accuracy while effectively preserving the semantic features, material segmentation, and per-point physical property distribution.

\noindent\textbf{Summary of contribution:} This paper makes the following contributions:
\begin{itemize}[leftmargin=10pt]
    \item We propose a framework that minimizes the latency of vision-guided physical property reasoning through both algorithmic and systematic optimization, thereby enabling  augmented visual cognition.
    \item We design a sensor-in-the-loop component that leverages real-time gaze tracking to infer a user's intention to locate the targeting object in cluttered environments. We further integrate this design into our optimization pipeline and conduct a real-world case study with Meta Aria Glasses.
    \item We benchmark our system on a gold-standard dataset against two SOTA baselines. The results show that our system achieves a significant latency reduction, demonstrating a performance gain of 62.9$\times$ over NeRF2Physics and 287.2$\times$ over PUGS on latency reduction.
\end{itemize}

\section{Vision-Guided Object's Physical Property Reasoning: Issues and Challenges}
\label{s:prelim}

In this section, we analyze current vision-guided object's physical property reasoning techniques, discussing their respective strengths and weaknesses.

\subsection{Query VLM with a Single Image}
\label{ss:query_image}

With the rapid advancement of Vision-Language Models (VLMs), a common and intuitive approach for inferring an object's physical properties is to directly query the VLM model using an image of the targeting object. 
While such methods can provide a fast estimate of physical properties, the results are often highly inaccurate due to three reasons. 

\noindent $\bullet$ \textbf{Lack of Absolute Size and Scale}. A single 2D image lacks the crucial geometric context and metric scale needed for accurate physical reasoning. Without this information, VLMs cannot distinguish between a miniature toy car and a life-size vehicle, leading to weight and mass predictions that can be off by orders of magnitude. 

\noindent $\bullet$ \textbf{Incomplete Visual Information}. A single viewpoint often fails to capture the entire object, leaving parts of it occluded, shaded, or out of frame. For example, a chair made of both metal and wood might be identified only as "metal" if the wooden legs are in shadow or blocked from view.

\noindent $\bullet$ \textbf{Absence of Explicit Spatial Expression}. 
While VLMs can estimate object-level physical properties such as mass, they are unable to provide continuous spatial predictions of voxel-level properties (e.g., density, hardness, or elasticity). 
For instance, while a VLM might infer an object is "glass", it cannot provide the specific elasticity or hardness at every individual point. Such fine-grained property maps are essential for applications in AR, where accurate physical simulation and precise object interaction are crucial.

Furthermore, directly querying a VLM for an object's physical properties with a single image lacks robustness in complex, real-world settings because the VLM's feature fusion pipelines, which are typically trained on controlled or synthetic datasets, often fail in cluttered environments where objects are occluded or overlap. This frequently leads to unstable and contradictory material assignments, making the system unreliable for practical applications. These limitations render single-image-based VLM query less appealing to real-time visual cognition applications.

\begin{figure}[t!]
    \centering
    \includegraphics[width=1\linewidth]{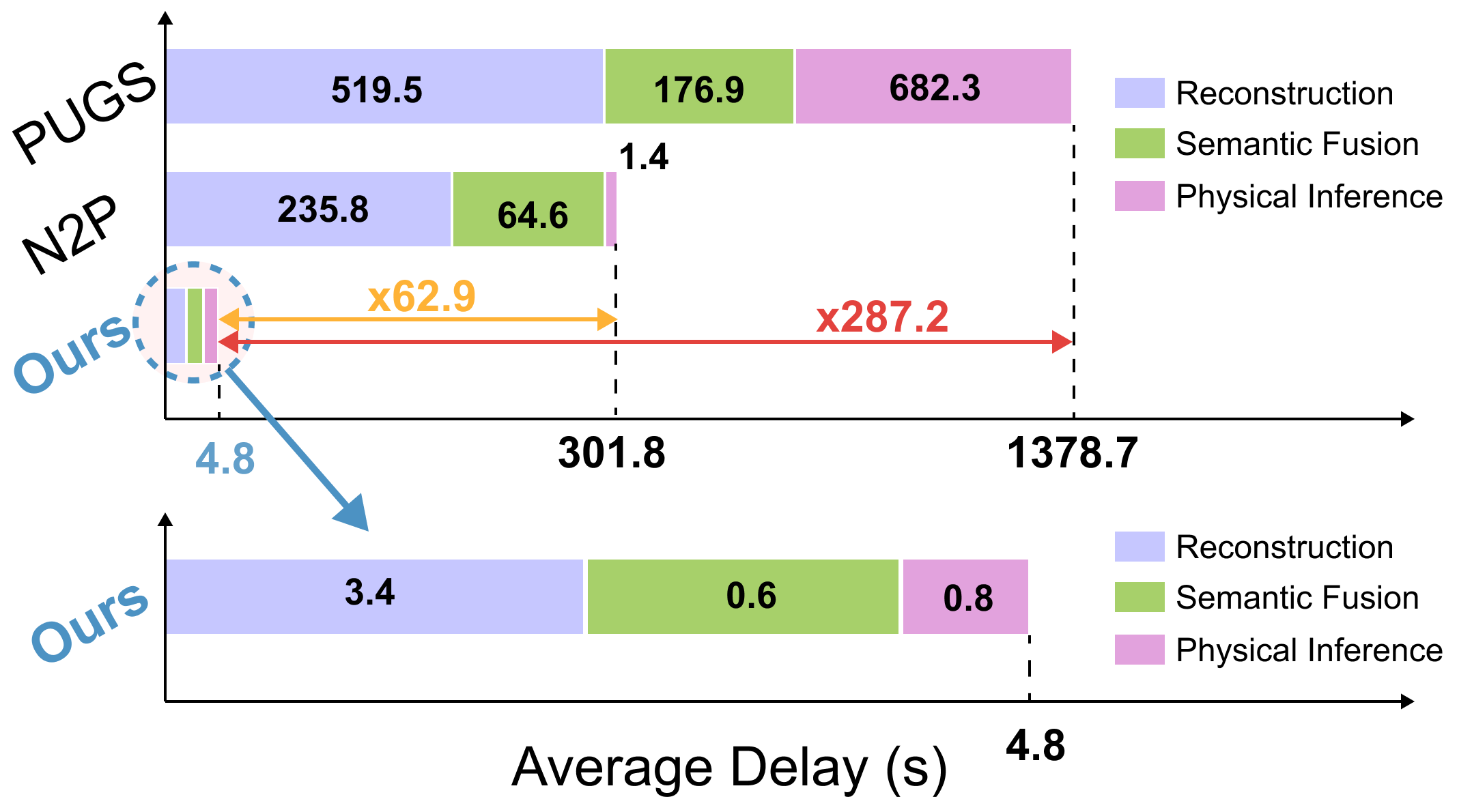}
    \vspace{-8mm}
    \caption{End-to-end system latency comparison of \sysname with two SoTA solutions NeRF2Physics (N2P)~\cite{zhai_physical_2024} and PUGS~\cite{shuai_pugs_2025}. \textnormal{By optimizing each processing component, \sysname achieves a significant latency reduction, demonstrating a performance gain of 62.9$\times$ over NeRF2Physics and 287.2$\times$ over PUGS.}}\vspace{-5mm}
    \label{fig:wholepip_delay}
\end{figure}

\begin{figure*}[t!]
    \centering
    \includegraphics[width=\linewidth]{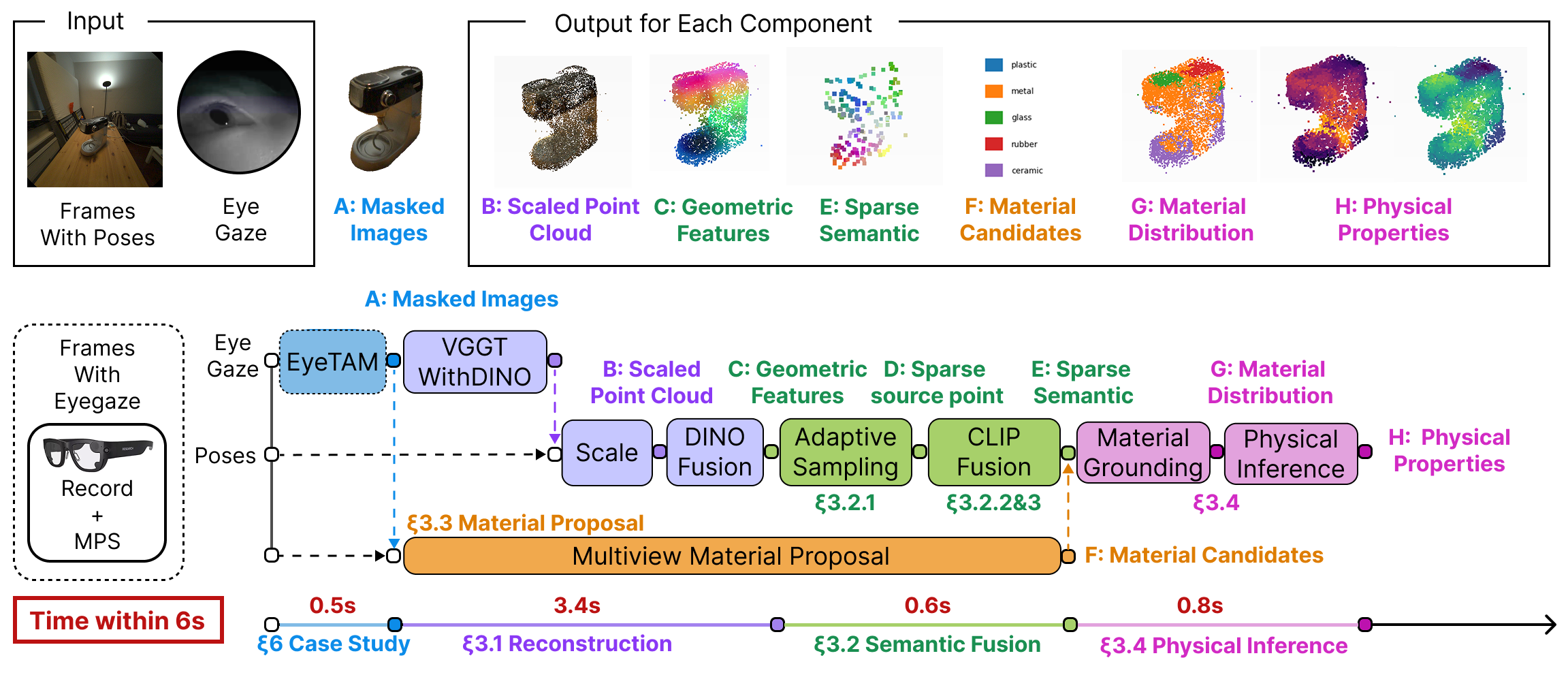}\vspace{-5mm}
    \caption{System workflow. \textnormal{\sysname accelerates multi-view physical property reasoning by (1) rapidly reconstructing object geometry, (2) fusing semantic cues from multi-view images into a unified 3D representation, and (3) mapping them into physical property fields for real-time reasoning.}}\vspace{-5mm}
    \label{fig:pipeline}
\end{figure*}

\subsection{Multi-View Object's Physical Property Reasoning: Promise and Pitfalls}
\label{ss:3Dreasoning}

Recent advances combining vision language models with neural 3D reconstruction~\cite{zhai_physical_2024, shuai_pugs_2025} have established a canonical pipeline for physical property reasoning. 
Despite variations in the specific algorithms used, these systems typically follow the following five key steps to answer what the object is, what its parts are made of, and where those parts are located, as elaborated below.

(1)~\textbf{3D Representation Generation}: a dense point cloud is reconstructed from a series of multi-view images to serve as the foundational 3D model of the object. 

(2)~\textbf{Semantic Feature Fusion}: these 3D points within the point cloud are then projected onto corresponding 2D views, where they are fused with visual semantic embeddings (e.g., "table leg" and "table panel" derived from pretrained VLMs such as CLIP) to create rich, point-level descriptors. 

(3)~\textbf{Material Proposal}: a large language model or a vision language model is then leveraged to generate candidate materials (e.g., metal, glass) for each object's component. 

(4)~\textbf{Semantic Material Grounding}: next, text embeddings of these candidate materials are compared with the fused visual features to produce a probabilistic distribution representing the likelihood of different materials. 

(5)~\textbf{Physical Property Inference}: finally, this material distribution is interpolated across the entire 3D representation to form a continuous property field. This field facilitates two distinct forms of inference: field-based estimation for localized properties (e.g., density, hardness) and integral estimation for global properties (e.g., total mass, center of mass).

By reconstructing a dense 3D point cloud from multiple viewpoints, multi-view physical property reasoning methods overcome the fundamental limitations of single-image approaches, such as the lack of absolute scale and occluded visual information. This makes them particularly ideal for augmented visual cognition applications, where a mobile user can naturally move around an object to capture images from various angles. This multi-view fusion also enables more accurate and robust estimations of properties like mass and volume. Furthermore, the fusion of features from multiple 2D images with a 3D representation allows the model to build a more comprehensive and stable understanding of an object's materials, making the system more reliable in complex, cluttered environments.

\noindent\textbf{Prohibitive Run-time Latency}. While effective, this multi-view pipeline suffers from significant processing delays, which creates a strong barrier to augmenting human visual cognition in real-time, interactive scenarios. We analyze this run-time processing delay in below.

As shown in Fig.~\ref{fig:wholepip_delay}, \emph{3D Representation Generation} (denoted as reconstruction in the figure) takes significant amount of time in multi-view physical reasoning pipeline. For instance, NeRF2Physics~\cite{zhai_physical_2024} requires 8–12 minutes of NeRF optimization, while PUGS~\cite{shuai_pugs_2025} spends 4–5 minutes training a 3D Gaussian Splatting model. This prohibitive delay is inherent to the iterative training and optimization processes of these 3D reconstruction models.
The \emph{Semantic Feature Fusion} (denoted as semantic fusion in the figure) adds extra processing overhead since thousands of 3D points must be projected into multiple views and passed through heavy vision–language encoders such as CLIP, often taking 1-14 minutes. 

In contrast, \emph{Material Proposal} is relatively quick, adding only a few seconds of latency due to LLM/VLM query time. Since this step solely requires the object's 2D images, its runtime can be effectively hidden by executing it in parallel with other stages of the pipeline.
However, the final stages, \emph{Semantic Material Grounding} and \emph{Physical Property Inference} (denoted as physical inference in the figure), also contribute a noticeable delay. For instance,  NeRF2Physics relies on $k$-NN interpolation (fast but noisy), while PUGS performs extra segmentation and contrastive learning that further increase the runtime by more than 10 minutes. Altogether, the end-to-end latency of existing multi-view physical reasoning pipeline is typically around 10 to 15 minutes for NeRF2Physics and 15 to 25 minutes for PUGS. This delay is far too long for mobile users who expect instant feedback from augmented visual cognition services, which is the core challenge that our system aims to overcome. 

\section{System Design}
\label{s:system_overview}

Our analysis in Section~\ref{ss:3Dreasoning} reveals that the primary bottlenecks in multi-view physical reasoning pipelines arise from three sources: (i) slow optimization-based 3D reconstruction, (ii) redundant semantic feature fusion, and (iii) complex post-processing techniques.  To overcome these challenges, we carefully analyze each stage of the pipeline and propose lightweight alternative solutions. 

Fig.~\ref{fig:pipeline} shows \sysname's workflow. We introduce three main design components, including (1) \textit{Rapid Geometric 3D Reconstruction} (\S\ref{sec:reconstruction}), which leverages pretrained end-to-end vision models to bypass iterative, photo-realistic training; (2) \textit{Efficient Semantic Feature Fusion} (\S\ref{ss:fuse}), which reuses intermediate features extracted from this vision model to align semantic features with non-structural 3D point cloud at the minimal latency; and (3) \textit{Fast Physical Property Inference} (\S\ref{ss:material_proposal}--\S\ref{ss:ppi}), which parallelizes postprocessing to minimize the latency of physical property field generation. 

We detail each of these components in turn.

\subsection{Rapid Geometric 3D Reconstruction}\label{sec:reconstruction}

Existing multi-view physical reasoning pipelines such as NeRF2Physics \cite{zhai_physical_2024} and PUGS \cite{shuai_pugs_2025} rely on training-based rendering methods like NeRF and 3D Gaussian Splatting to obtain the object's 3D point cloud. However, these rendering techniques by nature are designed to generate a photorealistic 3D representation that looks real from any angle, not just for generating a point cloud. This focus on visual perfection demands intensive training and optimization.

Our key insight is that this computationally expensive step is unnecessary for object physical property reasoning. Instead of a full-fledged reconstruction, we can directly leverage those end2end, one-time point cloud generation models to obtain an object's 3D point cloud to serve as the geometric foundation for physical reasoning, and then map material proposals directly obtained from the object's 2D images back to the 3D structure.
This would allow us to propagate the material information throughout the 3D representation, forming a physical property field without the need to generate a photorealistic yet time-consuming 3D object representation.

Based on this insight, we leverage VGGT~\cite{wang2025vggt}, a large pretrained, end2end vision model, to obtain the object's 3D point. Given just a few images of an object from different viewpoints\footnote{We ablate the impact of the number of images in evaluation.}, VGGT is able to produce a high-accuracy 3D point cloud, along with depth maps and camera poses, in a single forward pass within seconds. 
While using VGGT can drastically cut the 3D reconstruction time from 10 minutes to just two seconds, its thin, unified point cloud output lacks the necessary segmentation to differentiate between individual object components (\eg, "table leg", "panel"). Consequently, when we attempt to map semantic features from 2D images back to this point cloud (\S\ref{ss:fuse}), there is no structural information to guide the feature mapping and calibration process.

\begin{figure}
    \centering
    \includegraphics[width=0.9\linewidth]{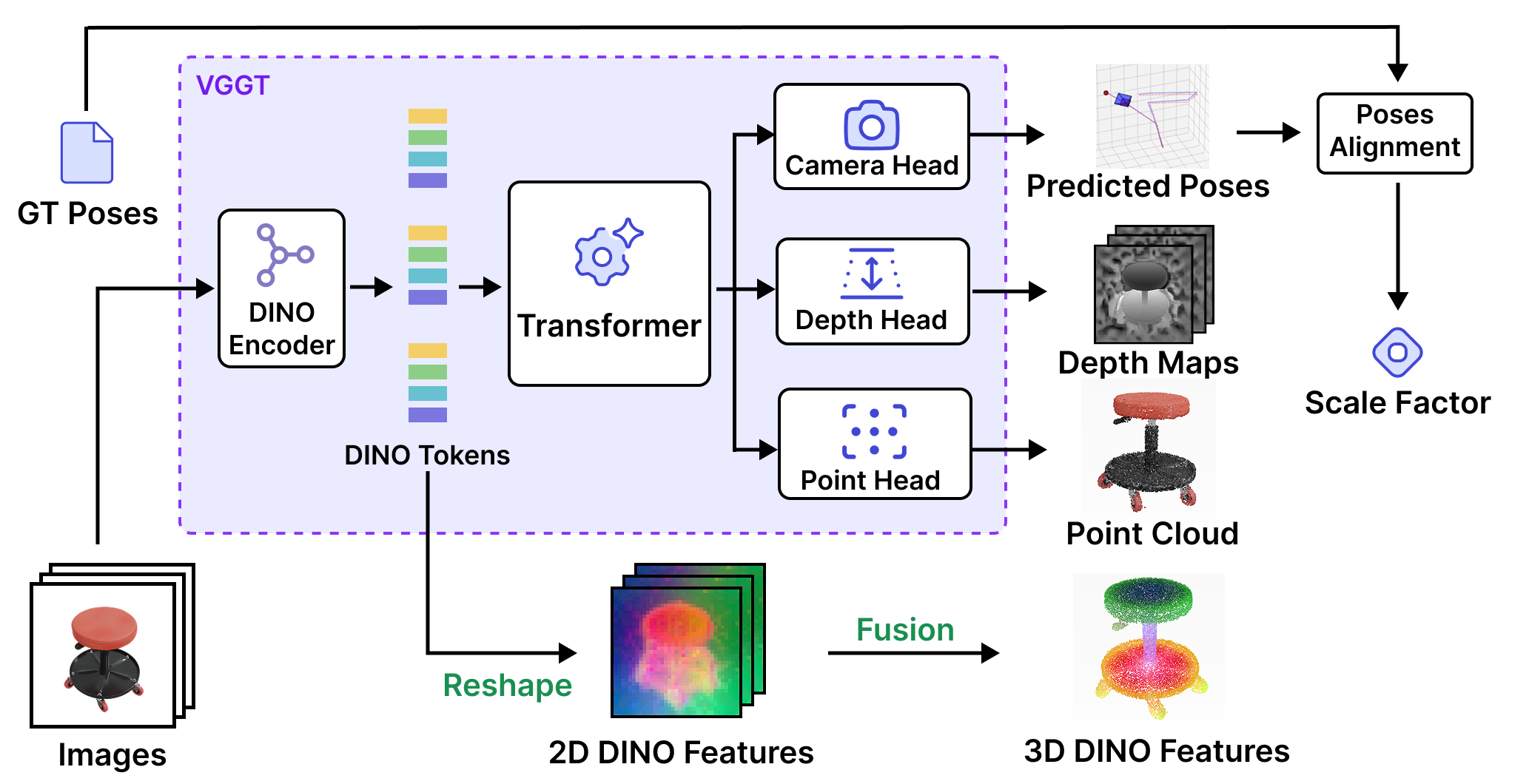}
    \vspace{-5mm}
    \caption{Rapid geometric 3D reconstruction. \textnormal{Given a small set of multi-view images, VGGT predicts camera poses, depth maps, and a coarse point cloud in a single forward pass. Intermediate DINO tokens are extracted and reshaped into 2D feature maps, which are then fused into a 3D semantic representation. A lightweight Sim(3) alignment step refines the predicted poses against ground-truth or auxiliary stereo poses to recover a consistent metric scale. }}
    \label{fig:ReconSys}
    \vspace{-4mm}
\end{figure}

\begin{table*}[t!]
\centering
\caption{The computation workload reduction of each design component in semantic feature fusion.
{\normalfont AS stands for Adaptive Sampling. VIF stands for View Importance Filtering. MFF stands for Multi-scale Feature Fusion. Counted in 5 randomly selected scenes. \sysname reduces the number of patches for processing from 28,040 to 738, achieving 38$\times$ reduction.}
}
\vspace{-3mm}\small
\begin{tabular}{l|cccc}
\hline
\hline
 & Baseline & AS (\S\ref{sss:ada-samp}) & AS + VIF (\S\ref{sss:vif}) & AS + VIF + MFF (\S\ref{sss:mff})\\
 \hline
 \textbf{Average number of embedded source points} & 3,505  & 82  & 82  & 82    \\
\textbf{Average patch number for each point} & 8  & 8  & 3  & 9 \\
\textbf{Number of embedded patches} & 28,040  & 656  & 246  & 738 \\
\hline
\hline
\end{tabular}\vspace{-5mm}
\label{t:delay_reduction_semantic_feature_fusion}
\end{table*}

To address this challenge, we leverage the DINOv2~\cite{oquab2023dinov2} features that VGGT already generates internally as part of its standard operation. These features are highly effective at capturing fine-grained geometric structure, enabling them to distinguish between individual object components and structural patterns. We extract these rich intermediate DINOv2 tokens by inserting forward hooks into VGGT's alternate attention layers. The tokens are then reshaped into 2D feature maps, which are aligned with the resolution of the input images and later reused for subsequent steps, such as adaptive sampling (\S\ref{sss:ada-samp}) and view importance filtering (\S\ref{sss:vif}). Notably, extracting DINO features introduces no extra latency, as these features are a natural byproduct of VGGT's single forward pass.
Fig. \ref{fig:ReconSys} shows the rapid geometric 3D reconstruction workflow.

Formally, we define the feature extraction as follows:
\[
\mathrm{Ply}_v,\; \mathrm{Depth}_v,\; F_v = \mathrm{VGGT}(I_v), \quad v \in V
\]
where $I_v$ denotes the $v$-th input image view, $\mathrm{Ply}_v$ is the predicted point cloud in camera coordinates, $\mathrm{Depth}_v$ is the corresponding depth map, and $F_v$ is the DINOv2 feature map aligned with the input resolution. We then fuse these multi-view 2D DINOv2 features into a unified 3D geometric representation for each point in the point cloud. This fusion process consists of two main steps: {\it visibility checking} and {\it feature aggregation}.

\noindent $\bullet$ \textbf{Visibility Checking}. For each 3D point $x_i$, we first project it onto all image views $v \in V$. A point is considered visible in a view $v$ only if its projected depth is less than or equal to the depth value predicted by VGGT at that pixel location, as determined by the visibility mask $M_{\mathrm{vis}}$:
\begin{equation}\footnotesize
M_{\mathrm{vis}}(x_i, v) =
\begin{cases}
1, & \text{if } \mathrm{proj\_depth}(x_i, v) \leq \mathrm{Depth}_v(\pi(x_i, v)) \\
0, & \text{otherwise}
\end{cases}
\end{equation}
where $\pi(x_i, v)$ is the projection of point $x_i$ onto view $v$.

\noindent $\bullet$ \textbf{Feature Aggregation}. Next, we aggregate the final feature for each 3D point by sampling and averaging the feature vectors from all its visible views:
\begin{equation}\footnotesize
\mathcal{F}_{\text{DINO}}(x_i) = \frac{\sum_{v \in V} M_{\mathrm{vis}}(x_i, v) \cdot F_v(\pi(x_i, v))}{\sum_{v \in V} M_{\mathrm{vis}}(x_i, v)}
\end{equation}
Once the per-point 3D features are constructed, a challenge remains before they can be used for physical property estimation is the unknown scale. The point cloud generated by VGGT lacks a real-world metric scale. Since quantities such as mass are directly dependent on an object's true size, resolving this ambiguity is essential. We address this by applying a Sim(3) similarity transformation~\cite{hartley2003multiple} to align the camera poses from VGGT with those from the device's stereo cameras (Fig.~\ref{fig:ReconSys}). This lightweight procedure yields a metrically-scaled point cloud.

The benchmark (\S\ref{ss:exp_latency_breadown}) shows that the average delay of this rapid geometric 3D reconstruction component is 3.378$s$, which is 69.8$\times$ and 153.7$\times$ faster than that in NeRF2Physics (235.8$s$), and PUGS (519.5$s$), respectively.

\vspace{-2mm}
\subsection{Efficient Semantic Feature Fusion}
\label{ss:fuse}

After obtaining the 3D point cloud, the next step is to ground semantic labels to the corresponding points on the point cloud (\ie, semantic feature fusion). With this, \sysname is able to tell us what part of the object each point represents (\eg, "table leg," "panel"). To achieve this goal, existing works~\cite{zhai_physical_2024,shuai_pugs_2025} follow a multi-step approach:

    (1)~\textbf{Step One: Subsampling the Point Cloud}: The original, dense point cloud is first thinned out to a smaller, more manageable set of key points. This reduces the number of calculations needed later.
    
    (2)~\textbf{Step Two: Finding Points on 2D Images}. For each photo of the object, every key 3D point on this thinned point cloud is projected onto the image to find its exact 2D location. If multiple points land on the same pixel, the point closest to the camera will be selected.
    
    (3)~\textbf{Step Three: Getting Descriptions from CLIP}:  A small image patch is then taken from around each of these 2D locations. This patch is fed into a powerful language-vision model like CLIP, which generates a descriptive vector (\aka, "feature vector") for that specific spot from that angle.
    
    (4)~\textbf{Step Four: Fusing Descriptions}:  Finally, the descriptive vectors from all the different photos are averaged together. The result is a single, rich semantic description for each key point in the 3D cloud.

Although this method can improve feature robustness, it introduces a significant runtime bottleneck: the frequent and numerous calls to the computationally expensive CLIP model. Specifically, to get detailed semantic coverage of an object's 3D point cloud, the system needs to process thousands of points. Since each of these points is seen from multiple images, the total number of patches to be extracted and encoded by the computationally expensive CLIP model becomes prohibitively high, leading to several minutes delay.

To minimize the delay on semantic feature fusion, we propose a three-stage method to minimize the number of patches to be processed by CLIP model while preserving the system robustness. We elaborate on each design below.

\subsubsection{\textbf{Adaptive Sampling}\label{sss:ada-samp}}

Current works adopt a standard voxel downsampling method that partitions the space into a uniform grid, which causes the number of sampled points to increase with object scale. This approach is suboptimal, as object scale does not necessarily correlate with structural complexity. To address this, we decouple sampling density from object scale by dynamically calculating the voxel size as follows:
\[
s_{voxel} = \sqrt[3]{\frac{V_{bound}}{N_{target}}}
\]
where $s_{voxel}$ is the adaptive voxel grid size, $V_{bound}$ and $N_{target}$ are the volume of boundary box for the point cloud and the desired number of points after downsampling, respectively. We determine $N_{target}$ based on the observation that most objects comprise fewer than 30 semantic components. To ensure a robust representation for each component, we allocate 4-5 sample points per part. Thus, we set $N_{target}\approx150$, a number sufficient for most common objects (though the actual number of sampled points may be slightly lower than $N_{target}$ due to the voxelization process), while allowing for reduction in simpler cases. 

Such a sparse sampling is effective because these source points, equipped with their powerful DINOv2 features from the reconstruction stage, are sufficient to represent the entire object's semantics. The rationale stems from a key property of these features: points on the same object component, even if spatially distant, share similar geometric characteristics and thus form distinct feature clusters. Consequently, this small set of points can act as strong semantic anchors. As we detailed later, these anchors guide a subsequent densification process to recover the detailed semantic map of the object.

As shown in Table~\ref{t:delay_reduction_semantic_feature_fusion},
in this way, we effectively reduce the averaged number of CLIP-embedded points from 3,505 to 82, which translates into 42$\times$ computation workload reduction compared against the baseline, while maintaining semantic diversity and structural coverage.

\subsubsection{\textbf{View Importance-based Filtering}}
\label{sss:vif}

\begin{figure}
    \centering
    \includegraphics[width=1\linewidth]{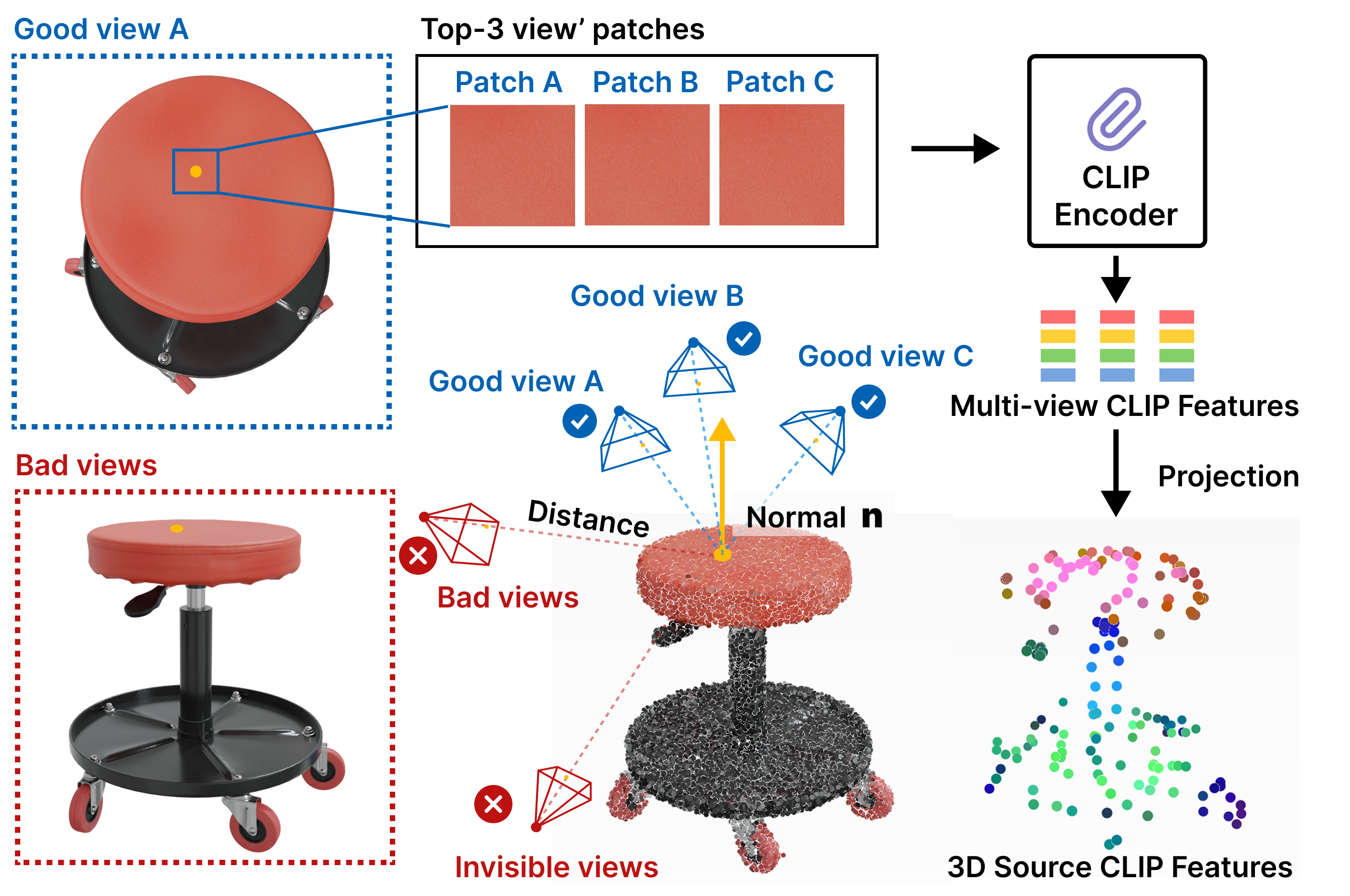}\vspace{-5mm}
    \caption{An illustration of view importance-based filtering.
    {\normalfont \sysname excludes those bad views that contribute less to semantic fusion, thereby reducing the runtime latency.}}
    \label{fig:view_selections}
    \vspace{-7mm}
\end{figure}
In practice, not all visible views contribute equally to the quality of the fused semantic feature. 
For instance, a clear, well-lit view of an object's handle contributes more useful information for semantic inference than a blurry, occluded view of the same object's body.
Hence, we can reduce the views to reduce the latency. We identified two primary sources of degradation:

\begin{itemize}[leftmargin=10pt]
    \item \textbf{Distance-induced resolution loss}: Views captured from a larger distance tend to offer lower spatial resolution, reducing the fidelity of texture details.
    \item \textbf{Surface-view angle mismatch}: Even if a view is spatially close to the point, a large deviation between the view direction and the surface normal can cause distortions, such as blur or reflection noise.
\end{itemize}

To mitigate these effects, we introduce a normal-depth-aware patch filtering mechanism that quantifies the importance of each patch based on its distance to the camera and its alignment with the surface normal.
We begin by estimating normal vectors for the sampled sparse point cloud. For each visible point-view pair, we compute two scores:  

\textbf{1. The Distance Score $S_{\text{dist}}$}, higher for closer views, reflecting better texture resolution:
\[
S_{\text{dist}}(z) = \frac{1}{1 + z}
\]
where $z$ is the depth of the target point in the camera coordinate frame.

\textbf{2. The Normal Angel Score $S_{\text{angle}}$}, higher when the view direction is well aligned with the surface normal:
\[
S_{\text{angle}}(\mathbf{v}, \mathbf{n}) 
= \max \left( 0, \; \mathbf{v} \cdot (-\mathbf{n}) \right)
\]
Here, $v$ is the unit vector from the camera center to the point, and $n$ is the unit surface normal (oriented outward). The dot product captures angular alignment, assigning higher scores to more front-facing views.
The overall patch importance score is defined as the product of the two:
\[
S_{\text{total}} = S_{\text{dist}}\cdot S_{\text{angle}}
\]

For each point, we rank all visible views by this score and retain only the top-$k$ views for feature extraction. As shown in Table~\ref{t:delay_reduction_semantic_feature_fusion}, our benchmark shows that this filtering mechanism further reduces the number of CLIP-embedded patches by around 3.3$\times$ while simultaneously improving computational efficiency and feature robustness by excluding distant or poorly aligned views.

\subsubsection{\textbf{Multi-scale Feature Fusion}}
\label{sss:mff}

After identifying the top-$k$ views for each point, we extract CLIP features from local image patches centered at the projected locations. However, relying on a single patch scale may fail to capture diverse object characteristics: small patches can miss larger geometric structures, while large patches may oversmooth local textures or introduce background noise.

To address this, following the method in LeRF~\cite{kerr_lerf_2023}, we adopt a multi-scale feature aggregation strategy. For each visible view, we extract patches at multiple spatial scales around the projected pixel and encode them using the CLIP image encoder. The resulting feature vectors are then averaged to form a multi-scale representation for each point. Finally, features from the top-$k$ views are fused via averaging to obtain the final semantic embedding for each point in the sampled point cloud. We formulate this process below:
\begin{equation}\footnotesize
\mathbf{f}(s) = 
\frac{1}{k} \sum_{v=1}^{k} 
\left(
    \frac{1}{|\mathcal{L}|} \sum_{l \in \mathcal{L}} 
    \phi\left( \mathcal{P}_{v}^{(l)}(s) \right)
\right)
\end{equation}
where \( \phi(\cdot) \) is the CLIP image encoder, 
\( \mathcal{P}_{v}^{(l)}(s) \) denotes a patch of scale \( l \) from view \( v \) centered at the projection of point \( s \), 
\( \mathcal{L} \) is the set of patch scales, 
and \( k \) is the number of selected views.
This hierarchical fusion captures both local texture details and broader geometric context, making the final representation more robust to object scale variation and viewpoint changes. While this approach processes an increased number of patches, the total remains manageable. As shown in Table~\ref{t:delay_reduction_semantic_feature_fusion}, the number of patches grows to only 738, which is still 38$\times$ lower than the baseline method.

\noindent\textbf{Parallelizing Semantic Feature Fusion}. 
An intuitive feature fusion pipeline would rely on a prohibitively slow nested loop, first iterating through each source point and then, for each point, serially projecting it onto every view for visibility checks and patch extraction. 
To minimize the processing latency further, we implement a highly efficient parallel algorithm that operate on our sparsely sampled source points. 
This algorithm begins by projecting all source points onto all views in a single, vectorized operation to simultaneously compute their distance scores. We then use the depth maps from the initial reconstruction to generate a unified visibility mask and batch-compute a normal angle score from pre-calculated normals. Based on these parallelly-computed scores, we can select the optimal patches timely, which are then aggregated into a single global batch for unified processing by the CLIP encoder. This design transforms the task from a point-by-point, serial process into a set of global, parallel operations, thereby saving significant amount of processing time.

Collectively, our optimizations reduce the latency of CLIP feature extraction and fusion to just 0.02 seconds. The entire semantic fusion stage, including the initial adaptive sampling and normal computation, is completed in under 0.6 seconds (Fig.~\ref{fig:wholepip_delay}). This represents a significant acceleration, achieving a 107.7$\times$ speedup over NeRF2Physics (64.6$s$) and a 294.8$\times$ speedup over PUGS (176.9$s$).

\begin{figure}[t!]
    \centering
    \includegraphics[width=0.8\linewidth]{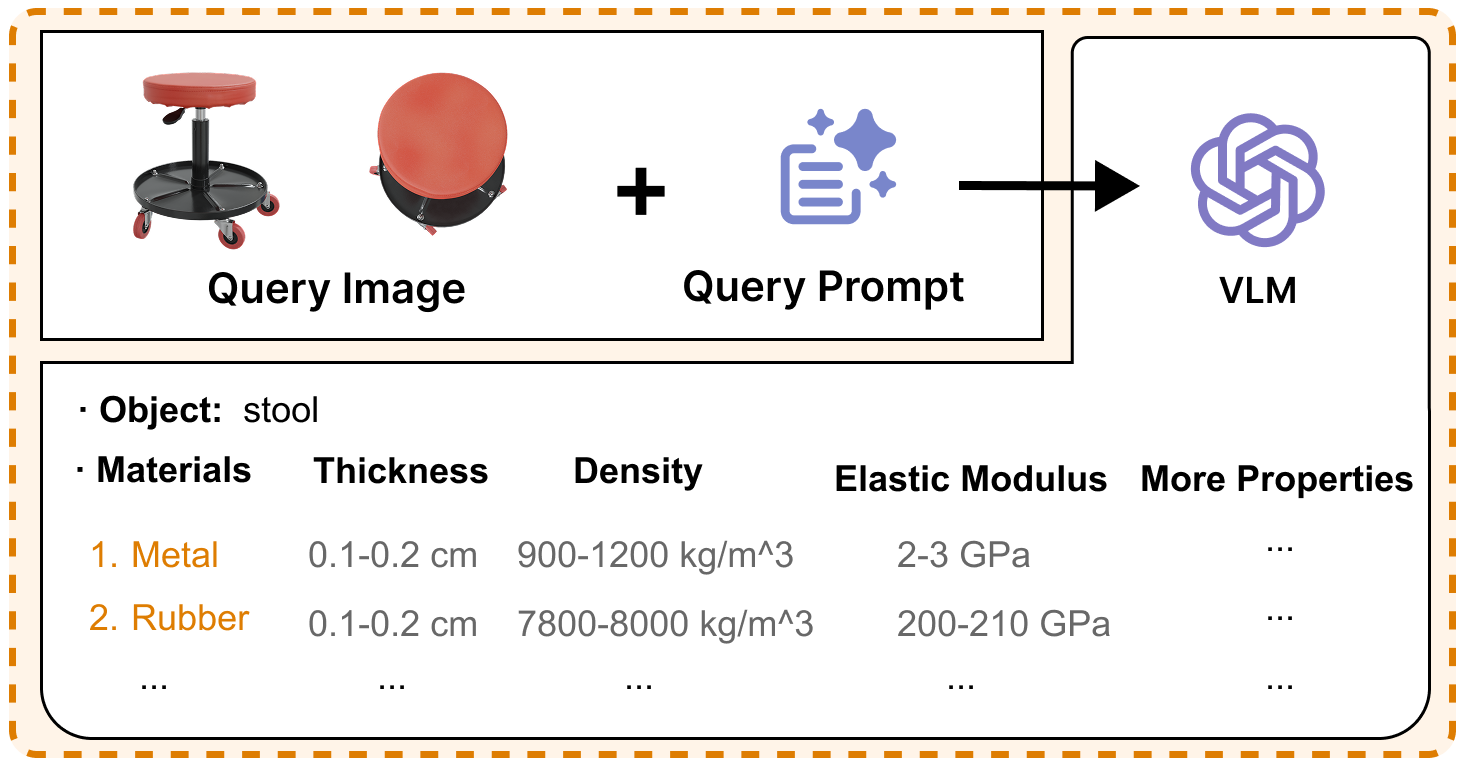}\vspace{-3mm}
    \caption{Material proposal pipeline.}\vspace{-7mm}
    \label{fig:material_proposal}
\end{figure}
\vspace{-3mm}
\subsection{Material Proposal}

\label{ss:material_proposal}

Once we populated the semantic features throughout the object's 3D point cloud, the next step is to generate material labels that match with these semantic features (\ie, CLIP features). 
As shown in Fig. \ref{fig:material_proposal}, we leverage GPT-4o to infer candidate materials and their associated physical properties, such as density, hardness, and elasticity. 

Unlike previous approaches~\cite{zhai_physical_2024,shuai_pugs_2025} that adopt a two-stage pipeline or on a single image---first performing image-to-text captioning, followed by text-to-property inference, which can lead to information loss. We directly prompt the VLM to perform zero-shot predictions based on selected images. 

Specifically, we randomly sample $2$--$4$ viewpoint images $I_m \in \mathcal{I}$ from our multi-view input and concatenate them into a single composite image, to avoid material omission caused by angle of view occlusion. This composite image is then provided to GPT-4o with tailored prompts to produce both a detailed semantic description and a set of candidate materials, together with their estimated thickness within the object and their corresponding physical property values. Formally, we obtain a material--property dictionary:
\[
    \mathcal{M} = \{(key_{k}, y_{k}, \theta_k)\}_{k=1}^{K},
\]
where $key_{k}$ denotes the candidate material name, $y_{k}$ denotes the associated property values or value ranges predicted by the model, and $\theta_{k}$ denotes the estimated thickness. 

This inference process typically requires up to 11 seconds on average depending on API latency and network conditions. 
Because this stage solely relies on the object's 2D images, we can run it as the user moves around to take photos of the object using her smart glasses. It can also run in parallel with both the geometric 3D reconstruction (\S\ref{sec:reconstruction}) and semantic feature fusion (\S\ref{ss:fuse}). This ensures that its processing time does not add to the end-to-end system latency.

\subsection{Physical Property Inference}
\label{ss:ppi}

With the material proposals obtained from vision language models and the semantic 3D point cloud on hand, we can further infer the object's physical properties.
An object's physical property can be broadly divided into two categories: voxel-level properties and object-level properties.
Voxel-level properties are characteristics of the individual volumetric pixels (voxels) that make up a 3D object's model, including object's material, hardness/density, and elasticity.
Object-level properties are characteristics of the object as a whole, often calculated by aggregating the voxel-level properties. 
They include mass, total weight, and center of mass, \etc
We describe how \sysname estimates each of these two categories.

\subsubsection{\textbf{Voxel-level Property Estimation}}
We first predict material distributions at each source point in the sampled point cloud using the CLIP features extracted via semantic fusion (\S\ref{ss:fuse}). Similar to NeRF2Physics~\cite{zhai_physical_2024} and PUGS~\cite{shuai_pugs_2025}, we apply a similarity-weighted kernel regression to infer the physical property value \( \rho(\mathbf{s}) \) at point \( \mathbf{s} \) as follows:
\begin{equation}\footnotesize
  \rho(\mathbf{s}) = \frac{\sum_{k=1}^{K}\exp(\omega_k[\mathbf{s}]/T) \cdot y_k}{\sum_{k=1}^{K}\exp(\omega_k[\mathbf{s}]/T)}  
\end{equation}
where \( y_k \) is the physical property (e.g., density) of material \( k \), \( \omega_k[\mathbf{s}] = \phi_{\text{CLIP}}(\mathbf{f}(\mathbf{s}), \textit{key}_k) \) is the CLIP similarity between the fused feature at point \( \mathbf{s} \) and the CLIP text embedding of material \( k \), and \( T \) is a temperature parameter controlling the confidence sharpness.

To obtain material predictions across the object's entire surface, we need to interpolate from the sparse source points $\{r_i\}_{i=1}^k$ to arbitrary query locations. However, direct interpolation based purely on Euclidean distance, as adopted by NeRF2Physics~\cite{zhai_physical_2024}, degrades significantly under sparse sampling. PUGS~\cite{shuai_pugs_2025} improves this by introducing a contrastively learned SAGA similarity, but still incurs over 10 minutes for its per-object contrastive learning process.

To address this, we propose a densification method guided by the rich geometric understanding of DINO features, which were pre-computed during the reconstruction stage (\S\ref{sec:reconstruction}). The core idea is to perform interpolation in the DINO feature space rather than Euclidean space. Since DINO features effectively capture an object's structural and component-level similarities, neighbors of a query point in this feature space are highly likely to belong to the same semantic part.

\begin{figure}
    \centering
    \includegraphics[width=1\linewidth]{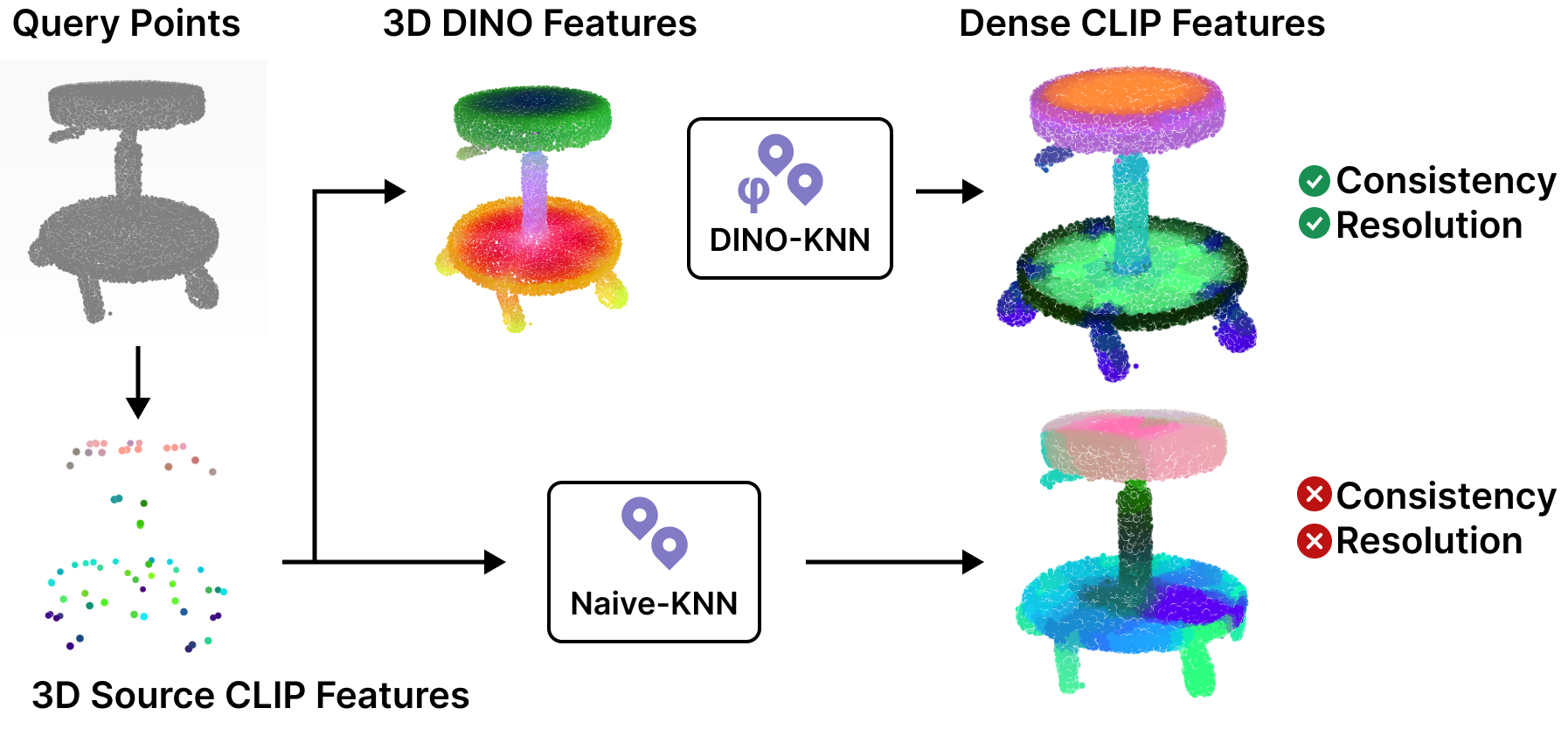}\vspace{-5mm}
    \caption{Comparison between DINO-KNN and Naive-KNN.
    {\normalfont We visualize the interpolation on CLIP features to show the difference. }}\vspace{-5mm}
    \label{fig:interpolation}
\end{figure}

Our method employs a k-nearest neighbor (k-NN) strategy based on DINO feature similarity. For any query point $q$ on the object's surface, we first identify its $k$ nearest source points, $\mathcal{N}_k(q)$, in the DINO feature space. To maximize efficiency, instead of interpolating high-dimensional CLIP features, we directly interpolate the pre-computed material probability vectors $\{\omega(r_j)\}_{r_j \in \mathcal{N}_k(q)}$. The interpolation uses a weighted average where weights are determined by the DINO similarity to the query point. The interpolated probability vector for the query point, $\omega(q)$, is formulated as:
\begin{equation}\footnotesize
    \omega(q) = \frac{\sum_{r_j \in \mathcal{N}_k(q)} \phi_{\text{DINO}}(\mathcal{F}(q), \mathcal{F}(r_j)) \cdot \omega(r_j)}{\sum_{r_j \in \mathcal{N}_k(q)} \phi_{\text{DINO}}(\mathcal{F}(q), \mathcal{F}(r_j))}
\end{equation}
where $\mathcal{F}(\cdot)$ extracts the DINO feature, $\omega(r_j)$ is the vector of material probabilities at source point $r_j$, and $\phi_{\text{DINO}}(\cdot, \cdot)$ is the cosine similarity. This new probability vector $\omega(q)$ is then used to compute the final physical property $\rho(q)$ via the kernel regression defined previously. By interpolating probabilities guided by geometric similarity, we generate dense predictions that are both fine-grained and structurally consistent as Fig \S\ref{fig:interpolation}.

This entire densification process, yielding properties like density and elastic modulus across the full surface, completes in approximately \textbf{0.8~s}. This runtime is comparable to the inference stage of prior works, but critically, our method \textbf{requires no expensive, per-object training or optimization phase}, unlike the 10-minute process required by PUGS.

\subsubsection{\textbf{Object-level Physical Property Estimation}}
\label{sss:object_level_physical_propert}

Similar to previous work, to estimate object-level properties like mass, we perform a voxel integration over volumes. Based on the voxel-level physical property, we treat each query point as a center of voxel with size $b\times b \times \theta$. $\theta$ is the estimated thickness of each material and the voxel surface size $b\times b$ is determined by the real scale and density of points.
Since the point cloud from VGGT is relatively uniformly distributed, we can assume that the dominant surface area of the object scales with the squared characteristic length $L$ of its bounding box. 
Given a point cloud with $N$ samples, the average surface area $\bar{A}$ represented by each point is approximated as:
\[
\bar{A} \approx \frac{(L \cdot \text{scale\_factor})^2}{N}.
\]
We then define the side length of this voxel surface as:
\[
b_{\mathrm{adap}} = \sqrt{\bar{A}}
\]
Then $b_{\mathrm{adap}}\times b_{\mathrm{adap}} \times \theta$ is a well-scaled voxel size for the following integration. 
\[\footnotesize
    \hat{\zeta} = \frac{\sum_{k=1}^{K}\exp(\omega_{k}[\mathbf{s}]/T) y_{k}{\theta_k}}{\sum_{k=1}^{K}\exp(\omega_{k}[\mathbf{s}]/T)}{b_{\mathrm{adap}}}^2\lambda
    ,
\]
where $\hat{\zeta}$ is the integrated physical property(\eg, mass), $\rho(\mathbf{s})$ is the interpolated physical property at point $\mathbf{s}$,$\lambda$ is an experience correction factor for thickness-based integration~\cite{zhai_physical_2024}.
$\omega_{k}[\mathbf{s}] = \phi_{\text{CLIP}}(\mathbf{f(s)}, key_{k})$.
This allows consistent estimation of aggregate properties over arbitrarily shaped objects.

\section{Implementation}
\label{s:implementation}

\begin{figure*}
    \centering
    \includegraphics[width=1\linewidth]{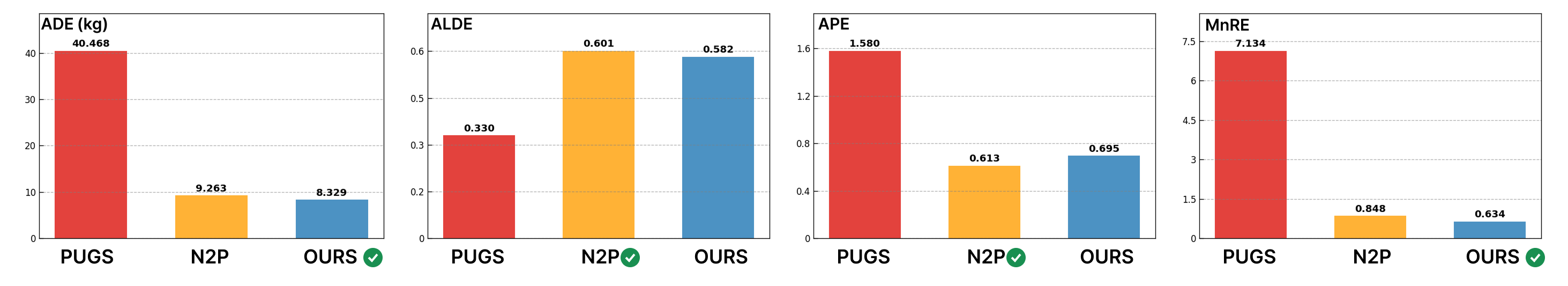}\vspace{-5mm}
    \caption{Mass Prediction Accuracy of \sysname (denoted as OURS), NeRF2Physics (denoted as N2P), and PUGS.}\vspace{-5mm}
    \label{fig:mass_bar}
\end{figure*}

\noindent\textbf{Software.} We implement our system in Python using the PyTorch framework. Our codebase extends the open-source repositories of PUGS~\cite{shuai_pugs_2025} and NeRF2physics~\cite{zhai_physical_2024}, for which we developed several novel modules. Furthermore, we modify the VGGT~\cite{wang2025vggt} model to expose its internal DINO tokens for our reconstruction process. For semantic understanding, we employ the CLIP ViT-L/14 backbone~\cite{wu_how_2022} via the OpenCLIP framework~\cite{ilharco_gabriel_2021_5143773} to extract feature embeddings. These are grounded by open-vocabulary material proposals from the OpenAI GPT-4o API~\cite{openai_gpt4_2023}. All feature manipulation operations are fully vectorized to ensure high GPU utilization. We use Open3D for geometry processing and visualization.

\noindent\textbf{Hardware.} All experiments are conducted on a server equip\-ped with an NVIDIA A6000 GPU with 48\,GB of memory, which serves as our primary evaluation platform.

\noindent\textbf{Parameters.} For reconstruction, we use 30 views per object, consistent with prior works~\cite{shuai_pugs_2025, zhai_physical_2024}. During semantic fusion, we set the target number of source points $N_{\text{target}}=150$, select the top 3 views for each point, and employ three patch scales with resolutions of 20, 40, and 60 pixels. In the physical inference stage, we adopt the parameters $T=0.1$ and $\lambda=0.6$ from NeRF2physics.

\section{Evaluation}
\label{s:eval}

We describe the evaluation of \sysname in this section.

\begin{figure}[t!]
    \centering
    \includegraphics[height=4.5cm]{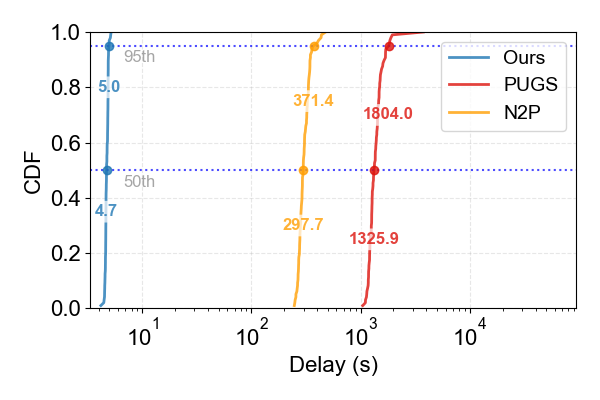}
    \caption{The CDF of the delay on geometric 3D reconstruction in \sysname, NeRF2Physics, and PUGS, respectively. 
    {\normalfont The 95th percentile of the delay is 4.7$s$, 297.7$s$, and 1325.9$s$ for \sysname, and NeRF2Physics, PUGS, respectively.}}
    \label{fig:delay_cdf}
\end{figure}

\subsection{Experiment Setups}
\textbf{Datasets.}  
We evaluate our system on the gold-standard ABO dataset~\cite{collins_abo_2022}, a collection of Amazon product listings that includes rich metadata, multi-view catalog images, and corresponding 3D models. The dataset also provides ground-truth physical properties such as material labels and object mass annotations across diverse product categories. Following prior work (NeRF2Physics and PUGS), we randomly sample 100 representative scenes from this corpus and group them into three categories according to object composition and environmental complexity for evaluation. Unless otherwise specified, we reconstruct each scenario using 30 views and perform subsequent physical inference. 

\noindent \textbf{Baselines.}  
We benchmark \sysname\ against two state-of-the-art physical reasoning pipelines:  
(1) \textit{NeRF2Physics}~\cite{zhai_physical_2024}, which leverages NeRF for geometry reconstruction and subsequent physical property inference.  
(2) \textit{PUGS}~\cite{shuai_pugs_2025}, a recent 3DGS-based framework that improves efficiency while retaining high-fidelity reconstruction and reasoning capabilities.    

\noindent \textbf{Metrics.}  
Following NeRF2Physics and PUGS, we primarily evaluate \textit{mass estimation accuracy} using four standard complementary error metrics:  
\textit{(i) ADE} (Absolute Deviation Error), the mean absolute difference between predicted and ground-truth mass values;  
\textit{(ii) ALDE} (Absolute Log-Deviation Error), the absolute difference between predicted and ground-truth values in the logarithmic domain, which reduces the bias from large-magnitude objects;  
\textit{(iii) APE} (Absolute Percentage Error), the absolute error normalized by ground-truth mass, highlighting relative deviations; and  
\textit{(iv) MnRE} (Mean Relative Error), the averaged relative error across objects. Note that for all metrics lower values indicate better accuracy, \textit{except ALDE where larger values reflect better alignment in the logarithmic space}. In addition to accuracy, we also measure system efficiency by reporting both the \textit{end-to-end latency} and the \textit{per-stage latency breakdown} of our pipeline.

\begin{table}[t]
\centering
\caption{The impact of scaling.}
\label{tab:scale_comparison}
\vspace{-3mm}\footnotesize
\begin{tabular}{l
                S[table-format=2.3] 
                S[table-format=1.3] 
                S[table-format=1.3] 
                S[table-format=1.3]}
\toprule\toprule
Object Size & {ADE $\downarrow$} & {ALDE $\downarrow$} & {APE $\downarrow$} & {MnRE $\uparrow$} \\
\midrule
\multicolumn{5}{l}{\textbf{Without-scaling}} \\
\hspace{3mm} Large   & 36.443 & 2.751 & 0.884 & 0.116 \\
\hspace{3mm} Middle  & 7.464  & 1.472 & 0.698 & 0.302 \\
\hspace{3mm} Small   & 1.717  & 0.927 & 1.212 & 0.536 \\
\textbf{\hspace{3mm} Overall} & \textbf{15.890} & \textbf{1.759} & \textbf{0.917} & \textbf{0.307} \\
\midrule
\multicolumn{5}{l}{\textbf{With-scaling}} \\
\hspace{3mm} Large   & 25.108 & 0.898 & 0.727 & 0.438 \\
\hspace{3mm} Middle  & 5.835  & 0.500 & 0.538 & 0.640 \\
\hspace{3mm} Small   & 0.698  & 0.594 & 0.526 & 0.597 \\
\textbf{\hspace{3mm} Overall} & \textbf{11.055} & \textbf{0.669} & \textbf{0.600} & \textbf{0.556} \\
\hspace{3mm} {\footnotesize {\textcolor{mygreen}{Improvement}}} 
  & {\footnotesize \textcolor{mygreen}{+30.4\%}} 
  & {\footnotesize \textcolor{mygreen}{+62.0\%}} 
  & {\footnotesize \textcolor{mygreen}{+34.6\%}} 
  & {\footnotesize \textcolor{mygreen}{+81.1\%}} \\
\bottomrule
\end{tabular}
\end{table}

\begin{figure*}
    \centering
    \includegraphics[width=1\linewidth]{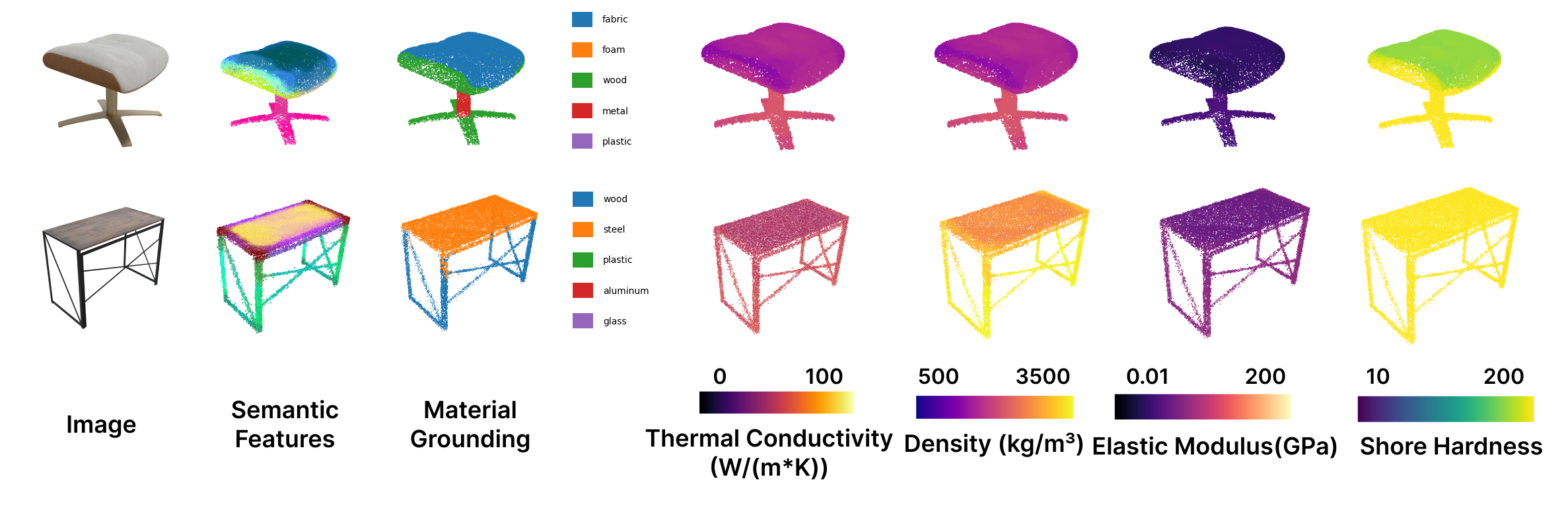}\vspace{-5mm}
\caption{Qualitative Results for Physical Properties 
{\normalfont Our system exhibits high fidelity in predicting per-point physical properties. The model correctly identifies material composition and maintains smooth, consistent property values across semantically coherent parts, such as the surface of the pan or the legs of the table.}}\vspace{-5mm}
    \label{fig:qualitative}
\end{figure*}

\vspace{-5mm}
\subsection{End-to-end System Performance}

\noindent \textbf{Latency.}  
Fig.~\ref{fig:delay_cdf} shows that \sysname\ reduces overall 3D physical inference time cost by \textit{two to three orders of magnitude}. The median latency is 4.7$s$, compared to 297.7$s$ for N2P and 1325.9$s$ for PUGS. This corresponds to a 63$\times$ speedup over N2P and a 282$\times$ speedup over PUGS. At the 95th percentile, \sysname\ completes in 5.0$s$, while N2P and PUGS require 371.4$s$ and 1804.0$s$, respectively. In practice, prior solutions incur delays of 10--25 minutes, whereas \sysname\ consistently runs in under 5 seconds, enabling real-time physical inference.  

\noindent \textbf{Accuracy.}  
As shown in Fig.~\ref{fig:mass_bar}, \sysname\ achieves mass prediction accuracy on par with N2P while clearly outperforming PUGS. Our ADE is 8.33\,kg versus 9.26\,kg (N2P) and 40.47\,kg (PUGS). On ALDE, we reach 0.582 compared to 0.330 (N2P) and 0.601 (PUGS). Similarly, \sysname achieves an APE of 0.695, close to N2P (0.613) and far below PUGS(1.580). For MnRE, \sysname achieves a score of 0.634, outperforming both N2P (0.848) and PUGS (\textit{7.134}). 

Overall, \sysname\ reduces latency by orders of magnitude while maintaining accuracy competitive with NeRF2Physics and substantially better than PUGS.  
This is enabled by our single forward VGGT-based geometry reconstruction and efficient feature fusion, which avoid the costly iterative optimization and numerous patch encoding used in NeRF and 3DGS pipelines.  

\subsection{Latency Breakdown}
\label{ss:exp_latency_breadown}
The main advantage of \sysname\ lies in its ability to drastically reduce the end-to-end latency of physical property reasoning.  
Table~\ref{tab:latency_categorized} (listed in Appendix) reports the runtime of each system component. The dominant cost is VGGT inference (3.43\,s), while subsequent modules such as semantic fusion (0.56\,s in total) and physical inference (0.80\,s) contribute comparatively little overhead. Scale computation and adaptive sampling are nearly negligible, each taking less than 20\,ms.

\subsection{Multiple Physics Properties Inference}
\label{sec:multi-physics}

Beyond mass estimation, \sysname\ can infer a spectrum of physical properties at fine spatial resolution. As shown in Fig.~\ref{fig:qualitative}, the system generates dense per-point maps of thermal conductivity, density, elastic modulus, and Shore hardness. These maps are spatially coherent and align with object structure—for instance, the cooking surface of a frying pan is identified as a high-conductivity metal region, while the legs of a wooden table exhibit higher stiffness and lower density. Such fine-grained physical distributions provide an interpretable view of an object’s material composition, going well beyond a single scalar weight value.  

Traditional vision–language models can at best assign coarse material categories such as \textit{metal}, \textit{plastic}, or \textit{wood}, but they cannot translate those labels into quantitative physical scales (e.g., conductivity in W/(mK) or stiffness in GPa). By combining CLIP embeddings with GPT-4o–guided material grounding, \sysname\ bridges this gap and outputs continuous, physically meaningful parameters directly from visual input. 

\vspace{-3mm}
\subsection{Micro-benchmarks}
\noindent $\bullet$ \textbf{Impact of Number of Views.}  
We vary the number of input views/images (5, 10, 30) to study its impact on reasoning latency. As shown in Table~\ref{t:impact_of_view_numbs}, the latency drops from 4.600$s$ to 2.214$s$ as we reduce the number of views (images) from 30 to 10. The delay further drops to 1.680$s$ as we further reduce the number of views to 5. Moreover, we found the major delay reduction comes from the 3D structual reconstruction.

\noindent $\bullet$ \textbf{Impact of Scaling.} 
\label{sec:scale_ablation}
We conduct an ablation study to validate the importance of real-world scaling in mass estimation. Table~\ref{tab:scale_comparison} shows that incorporating the scaling factor yields substantial accuracy gains across all object sizes. Overall, ADE improves by 30.4\%, ALDE by 62.0\%, APE by 34.6\%, and MnRE by 81.1\%. This significant improvement comes at a negligible computational overhead of only \textit{20\,ms}, confirming that real-world scaling is both a crucial and efficient component of our design.  
\begin{table}[t]
\centering
\caption{The delay(s) in different number of images used for physical property reasoning.}
\begin{adjustbox}{width=1\columnwidth}
\begin{tabular}{c|cccccc}
\hline
\hline
\textbf{Views} & \textbf{Reconstruction} & \textbf{Semantic Fusion} & \textbf{Physical Inference} & \textbf{Overall} \\
\hline
30 & 3.348 & 0.561 & 0.691 & 4.600 \\
10 & 0.937 & 0.588 & 0.689 & 2.214 \\
5  & 0.480 & 0.569 & 0.631 & 1.680 \\
\hline
\hline
\end{tabular}
\end{adjustbox}
\label{t:impact_of_view_numbs}
\end{table}

\section{Case Study with Smart Glasses}
\label{sec:realworld-eval}
To validate \sysname\ in a practical Augmented Reality (AR) setting, we conduct a case study using a real-world dataset of 15 objects captured with Meta Aria smart glasses~\cite{engel2023project} in a cluttered retail environment. For this case study, we developed \textbf{EyeTAM}, a plug-in module that integrates gaze tracking with EfficientTAM~\cite{kwon_efficient_2023} to automatically identify and segment the user's object of interest, demonstrating our system's robustness in the wild.

\noindent \textbf{Experimental Setup.}
We prototyped our system on Meta Aria Glasses (V1). Due to current firmware restrictions on on-device computation, we adopted a \textit{record-then-process} workflow: raw sensor streams were captured on the glasses and offloaded to an edge workstation equipped with an NVIDIA RTX A6000 GPU for processing. The Aria glasses provide synchronized feeds from dual SLAM cameras (30 FPS), a high-resolution RGB camera (15 FPS, 1408×1408), dual eye-tracking cameras (30 Hz), and IMUs. We use the Meta Aria SDK~\cite{engel2023project} to extract SLAM trajectories and camera poses for 3D reconstruction. Each capture session focused on a single object, yielding approximately 200 views. Our EyeTAM module processes the streams to extract 8 segmented images per object within 0.5s, and feed them to \sysname.

\begin{figure}[t!]
    \centering
    \includegraphics[width=1\linewidth]{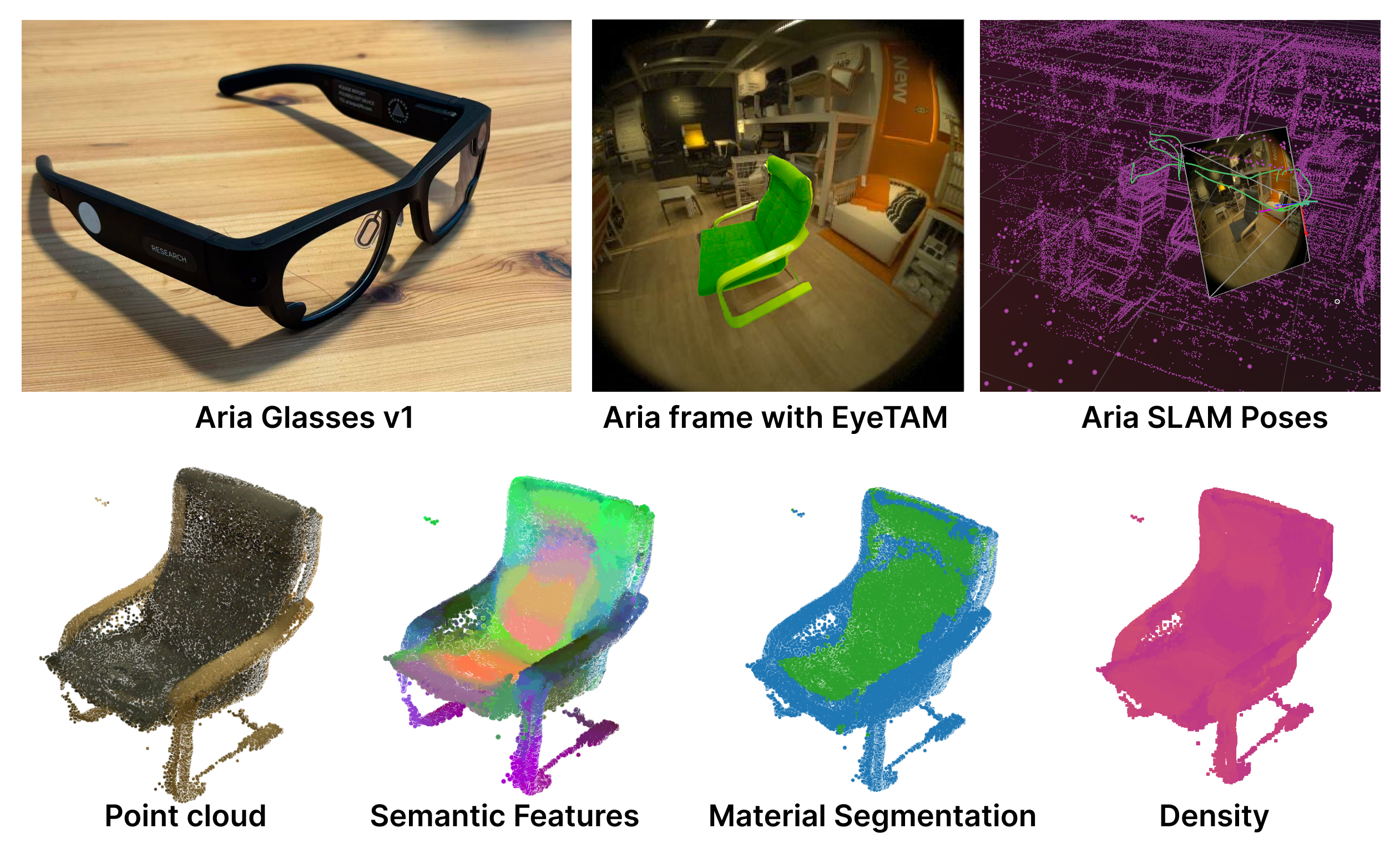}\vspace{-3mm}
    \caption{Material reasoning in real-world scenario.}
    \label{fig:Aria}
\end{figure}

\noindent \textbf{Results.}
Fig.~\ref{fig:Aria} qualitatively demonstrates our pipeline's output from an Aria recording. The process begins with the EyeTAM module tracking the user's target, followed by SLAM-based trajectory estimation and point cloud reconstruction. From this reconstruction, \sysname\ maps semantic features, performs material segmentation, and infers per-point physical property distributions, such as density. Quantitatively, the mass estimation results degrade only slightly compared to controlled captures, demonstrating that our system provides robust estimates even with challenging real-world data and fewer views (Table~\ref{tab:error_metrics_horizontal}). Such per-point predictions provide a richer physical understanding than a single scalar value, helping the system and ultimately the user interpret how different parts of an object contribute to its overall weight, stability, or fragility in real-world interactions.
\begin{table}[t]
    \centering
    \caption{Mass accuracy results on real world data. {\normalfont The ground truth values are from IKEA official website.}}
    \label{tab:error_metrics_horizontal}
    \begin{tabular}{cccc}
        \toprule
        \textbf{ADE} & \textbf{ALDE} & \textbf{APE} & \textbf{MnRE} \\
        \midrule
        5.851 & 1.175 & 1.133 & 0.447 \\
        \bottomrule
    \end{tabular}
\end{table}

\section{Related Works}
Our system relates to the augmented reality, vision language model, and egocentric HCI.

\noindent \textbf{(1) Physics in Augmented Reality.}
Early AR systems focused on overlaying semantic information such as labels or instructions without reasoning about underlying physics~\cite{dogan_augmented_2024,numan_spaceblender_2024}. More recent efforts combine neural scene representations with physics estimation: NeRF2Physics~\cite{zhai_physical_2024} integrates volumetric reconstruction with CLIP-guided material grounding to infer mass and friction, while PUGS~\cite{shuai_pugs_2025} leverages 3D Gaussian Splatting and contrastive learning to estimate part-level materials and aggregate properties. Parallel threads in robotics and computer vision studied intuitive physics from video sequences~\cite{wu2015galileo,battaglia2016interaction}, material recognition~\cite{groth2018shapestacks} and reflectance estimation~\cite{bell2013opensurfaces}, and visual weight~\cite{nyalala2021weight}. However, these approaches typically require extensive offline optimization or curated datasets, limiting their deployment in mobile AR. In contrast, our work introduces a \emph{real-time, marker-free system} that collapses minutes of optimization into seconds, enabling physical property inference directly from egocentric multi-view streams.

\noindent \textbf{(2) 3D Vision–Language Grounding.}
Linking natural language to 3D reconstructions has been accelerated by foundation models. On the 2D side, CLIP~\cite{radford2021learning}, BLIP~\cite{li2022blip}, Flamingo~\cite{alayrac2022flamingo}, and DINO/DINOv2~\cite{caron2021emerging,oquab2023dinov2} provide powerful image–text representations, while SAM~\cite{kirillov2023segment} enables promptable segmentation. Extending these to 3D, LERF projects CLIP features into NeRF for open-vocabulary queries~\cite{kerr_lerf_2023}, followed by works that align language with Gaussian splats~\cite{li_4d_2025,qin2024langsplat}, and meshes~\cite{hu2024sparselgs}. Open-vocabulary 3D segmentation and mapping frameworks~\cite{peng_openscene_2023} fuse cross-view features but suffer from high runtime costs. Advances in fast reconstruction, from instant-ngp~\cite{muller2022instant} to 3D Gaussian Splatting~\cite{kerbl20233d}, enable near real-time geometry but still incur overhead during semantic fusion. Our work differs by combining a \emph{one-shot reconstruction backbone with adaptive subsampling and multi-scale CLIP fusion}, thus achieving real-time grounding suitable for mobile AR deployment.

\noindent \textbf{(3) Egocentric XR Sensing and Interaction.}
Egocentric platforms such as Meta Aria~\cite{engel2023project} and Ego4D~\cite{grauman2022ego4d} provide synchronized multimodal datasets for embodied perception. Systems like SemanticAdapt and Teachable Reality emphasize semantic overlays, while collaborative XR spaces explore multi-user interaction~\cite{numan_spaceblender_2024}. Networking-focused systems such as Habitus~\cite{zhang2024habitus} improve XR delivery using multipath and pose-guided throughput prediction, yet they optimize communication rather than physical inference. Our work advances this line by \emph{fusing gaze signals with fast geometry and semantic grounding to deliver physical overlays}, enabling robust, real-time object-level physics reasoning in cluttered egocentric environments.

\section{Conclusion}
We have presented \sysname, a system designed for accelerating object's physical property reasoning for augmented visual cognition. By introducing a series of simple yet effective optimizations, \sysname reduces end-to-end latency from tens of minutes to under 6 seconds while maintaining high accuracy. Our work establishes a foundation for future augmented visual cognition, enabling XR applications that require accurate, low-latency reasoning about the physical world.

\bibliographystyle{ACM-Reference-Format}
\bibliography{sample-base, references,ref-lhb}

\appendix

\section{Delay Analysis}
\begin{table}[b!]
    \centering\vspace{-5mm}
    \caption{Latency breakdown.
    {\normalfont All times are reported in seconds. We did not count in the material proposal delay (\S\ref{ss:material_proposal}) since it can run in parallel with these modules and thus would not block each other.}}\vspace{-5mm}\footnotesize
    \label{tab:latency_categorized}
    \begin{tabularx}{\columnwidth}{l >{\RaggedLeft}X}
        \toprule
        \textbf{Component} & \textbf{Latency (s)} \\
        \midrule
        \multicolumn{2}{l}{\textbf{Geometric Reconstruction}} \\
        \hspace{1em}VGGT Inference & 3.378 \\
        \hspace{1em}DINO Aggregation  & 0.047 \\
        \hspace{1em}Scale Computation & 0.002 \\
        \multicolumn{2}{l}{\textbf{Semantic Fusion}} \\
        \hspace{1em}Normal Vector Computation & 0.508 \\
        \hspace{1em}Adaptive Sampling & 0.015 \\
        \hspace{1em}CLIP Feature Fusion & 0.035 \\
        \multicolumn{2}{l}{\textbf{Physical Inference}} \\
        \hspace{1em}Material Segmentation & 0.445 \\
        \hspace{1em}Property Inference & 0.355 \\
        \midrule
        \textbf{Total} & \textbf{4.785} \\
        \bottomrule
    \end{tabularx}
\end{table}

\end{document}